# Quantification of Tenseness in English and Japanese Tense-Lax Vowels: A Lagrangian Model with Indicator $\theta_1$ and Force of Tenseness Ftense(t)

## Tatsuya Ishizaki

*Utsunomiya Kyowa University, Utsunomiya, Tochigi, 320-0811, Japan*
(E-mail: t-ishizaki@kyowa-u.ac.jp, capestone0214@gmail.com)

## Abstract

The concept of vowel tenseness has traditionally been examined through the binary distinction of tense and lax vowels. However, no universally accepted quantitative definition of tenseness has been established in any language. Previous studies, including those by Jakobson, Fant, and Halle (1951) and Chomsky and Halle (1968), have explored the relationship between vowel tenseness and the vocal tract. Building on these foundations, Ishizaki (2019, 2022) proposed an indirect quantification of vowel tenseness using formant angles ($\theta_1$ and $\theta_{F1}$) and their first and second derivatives, $dZ_1(t)/dt = \lim \tan \theta_1(t)$ and $d^2Z_1(t)/dt^2 = d/dt \lim \tan \theta_1(t)$. The Japanese vowel /uR/ has been empirically demonstrated to exhibit tenseness, indicating the existence of a Tense/Lax pair. This study extends this approach by investigating the potential role of a force-related parameter in determining vowel quality. Specifically, we introduce a simplified model based on the Lagrangian equation to describe the dynamic interaction of the tongue and jaw within the oral cavity during the articulation of close vowels. This model provides a theoretical framework for estimating the forces involved in vowel production across different languages, offering new insights into the physical mechanisms underlying vowel articulation. The findings suggest that this force-based perspective warrants further exploration as a key factor in phonetic and phonological studies.

## 1. INTRODUCTION
### 1.1 Background

Regarding tenseness of vowels, Jakobson, Fant and Halle (1951) and Chomsky and Halle (1968) described the characteristics of tense vowels and lax vowels. They pointed out the relationship between tenseness and vocal tract. Specifically, they focused on the positional relationships of the vocal tract in examining vowel tenseness and argued that in tense vowels, the vocal tract deviated further from its neutral position compared to lax vowels. This suggested that considering the vocal tract, which constitutes the articulatory organs, was beneficial in understanding vowel tenseness.

Jakobson, Fant, and Halle (1951) discussed the tenseness of vowels and provided a detailed account of the correlation between tense and lax vowels. Building on this, Chomsky and Halle (1968) further examined the relationship between vowel tenseness and the vocal tract configuration. They focused on the spatial arrangement of the vocal tract, arguing that tense vowels deviate more from the neutral position than lax vowels. This suggests that considering the vocal tract, which constitutes the articulatory system, is useful for understanding vowel tenseness. In this way, vowel tenseness is clearly defined based on the relative positioning of the vocal tract.

On the other hand, to the best of our knowledge, no quantitative definition based on specific physical parameters has been established. For example, in English, vowels are distinguished as either tense or lax. Vowels classified as long vowels or diphthongs are generally recognized as tense vowels, whereas those categorized as short vowels are regarded as lax vowels. However, this distinction is not based on a quantitative definition. While this classification appears to rely on vowel duration, vowel duration itself varies due to contextual factors such as sentence environment and the presence or absence of stress. Therefore, it is possible that some parameter other than duration governs vowel tenseness.

Regarding the spatial configuration of the vocal tract, as discussed by Jakobson, Fant, and Halle (1951) and Chomsky and Halle (1968), previous studies, such as those by Chiba and Kajiyama (1941) and Fant (1960), have demonstrated that the vertical position of the tongue and jaw, which is one of the determining factors of vocal tract shape, correlates with the first formant frequency. These findings suggest a correlation between vowel tenseness, vocal tract configuration, and the first formant frequency—three key acoustic-phonetic properties. Building on this, Ishizaki (2018c) proposed the possibility of a quantitative analysis of the temporal dependency of vowel tenseness based on first formant frequency transitions over time.

Ishizaki (2018c) conceptualized vowel tenseness as a time-dependent property within a finite duration. In this study, we further develop the discussion, which was initially explored in Ishizaki (2019) (in Japanese), extending it beyond finite-duration vowel tenseness to examine vowel tenseness at a specific moment in time, $t$.

## 1.2. Acoustic Properties and Previous Research：Formant Movement

Formant movement refers to the variation in formant frequency over time. Until around the 1980s, research on formant frequencies was conducted without considering their temporal dependence. However, since the 1990s, advancements in speech analysis technology have led to increased attention to formant movement in the field of acoustic phonetics across various languages.

Figure 1 illustrates the acoustic properties commonly used in vowel quality research. The vertical axis represents formant frequency, while the horizontal axis represents time. The figure shows a vowel produced between the onset and offset of the time axis, with this time interval referred to as duration. A common analytical approach is to consider the relatively stable resonance frequency with minimal fluctuation during the vowel's duration as its formant frequency. In studies on formant movement using words in which vowels are preceded and followed by consonants, a frequent methodology is to define and analyze formant movement within the 33% to 66% time range. This approach aims to eliminate the influence of adjacent consonants on vowel quality, known as formant transition.

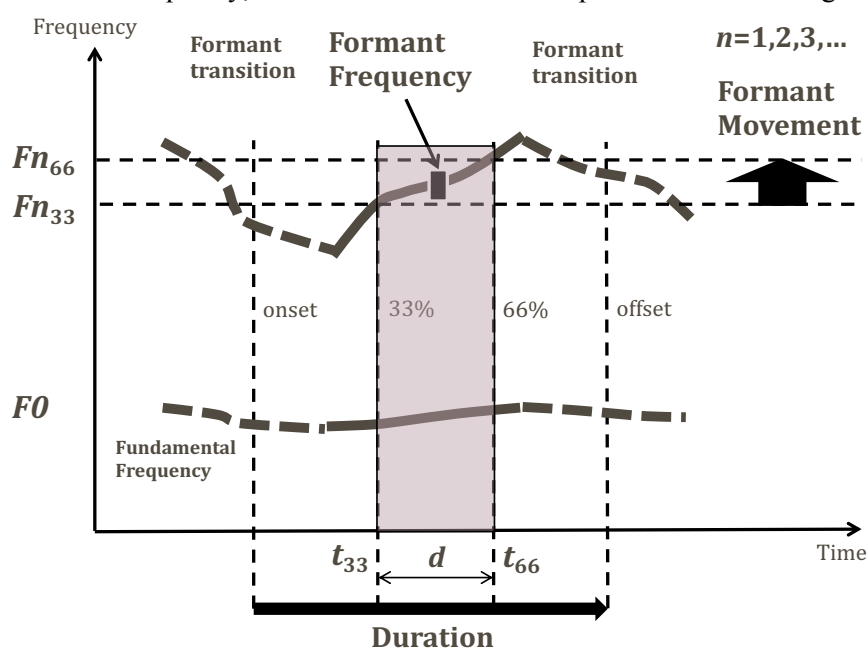

**Figure 1. Acoustic Properties Used in Vowel Quality Research**

Research on vowel formant movement spans a wide range of topics, including acoustic properties (Hillenbrand et al., 1995; Balbonas & Daunys, 2015), human vowel perception (Watson & Harrington, 1999; Iverson & Evans, 2007), automatic vowel recognition systems (Wrede et al., 2000), dialectal variation (Jacewicz & Fox, 2013), individual and age-related differences (McDougall & Nolan, 2007; Yang & Fox, 2013, 2017), tenseness (Ishizaki, 2018a, 2018b), pronunciation training (Schwartz, 2015; Ishizaki, 2015, 2016, 2018c, 2020, 2021, 2022, 2023), and lexical access models (Slifka, 2003).

## 2. Objectives and Methodology

This study explores the quantitative analysis of vowel tenseness. First, it presents a quantitative definition of tenseness and examines its relationship with the vocal tract space. Additionally, the study discusses the relative relationships of tenseness and considers an index for tenseness at a given time $t$. Based on these definitions, it quantitatively clarifies the tenseness of English tense /iː/ and lax /ɪ/, as well as English tense /uː/ and lax /ʊ/. These findings provide novel phonetic data. Furthermore, the study conducts a comparative analysis of the tenseness of Japanese tense /uR/ (HL) and lax /uR/ (LH) in relation to the corresponding English vowels. Through a physical analysis, it estimates the forces involved in the articulation of English tense /uː/ and Japanese tense /uR/ (HL). Finally, to estimate the forces exerted during vowel production, the study examines a simplified model using the generalized Lagrange equation, laying the groundwork for future expanded research.

The study analyzed vowel tenseness using 135 English speech samples from the Cambridge Dictionary, Oxford Learner's Dictionaries, and Longman English Dictionary. Measurements included tenseness parameters ($\theta_1$) and first formant frequency ($F1$) for tense /iː/ (67 samples, e.g., eat) and lax /ɪ/ (68 samples, e.g., it). Statistical analyses (t-tests, two-way ANOVA) were conducted using R.

For Japanese /uR/, 71 samples from Tohoku University-Matsushita Word Speech Database (TMW) were categorized as HL (34) and LH (37), with example words including Tesūryō (HL) and Sūgaku (LH). The English vowel data from the Cambridge Dictionary included 23 samples: tense /uː/ (9, e.g., boot) and lax /ʊ/ (14, e.g., foot).

Formant transitions (33%-66% of vowel duration) were analyzed using Praat to minimize contextual influence, providing a quantitative basis for comparing English and Japanese vowel tenseness. This study focuses on the articulation of close vowels in English and Japanese.

## 3. Definition of Vowel Tenseness
## 3.1 Time Dependence of Vowel Tenseness

Jakobson, Fant, and Halle (1951) provided a detailed analysis of the correlation between tense and lax vowels. They argued that "in a tense vowel the sum of the deviation of its formants from the neutral position is greater than that of the corresponding lax vowel" (Jakobson, Fant, and Halle, 1951, p. 36), suggesting that vowel tenseness can be distinguished by summing the deviations of formant frequencies. Furthermore, they stated that "the higher tension is associated with a greater deformation of the entire vocal tract from its neutral position" (Jakobson, Fant, and Halle, 1951, p. 38), highlighting the relationship between vowel tenseness and the displacement of the vocal tract.

Chomsky and Halle (1968) also examined the connection between vowel tenseness and vocal tract configuration. They noted that "one of the differences between tense and lax vowels is that the former are executed with a greater deviation from the neutral or rest position of the vocal tract than are the latter" (Chomsky and Halle, 1968, p. 324). This definition, based on the relative positioning of the vocal tract, underscores the importance of articulatory structures in understanding vowel tenseness.

Additionally, it is widely recognized that the first formant frequency (*F1*) correlates with changes in the spatial configuration of the vocal tract along the vertical axis.

Building on these phonetic findings, Ishizaki (2018c) proposed a definition of vowel tenseness, as presented in Table 1 and Figure 2. Given the difficulty of precisely quantifying tenseness at a single point in time, Ishizaki introduced the concept of time dependence in the analysis of vowel tenseness, using the angle of the first formant frequency ($\theta_1$) as an index. By employing $\theta_1$, vowel tenseness is not merely classified as either "tense" or "lax," but rather quantified as a continuous numerical value.

**Table 1. Definition of Vowel Tenseness**

| Vowel Tenseness | The temporal variation in vowel tenseness can be indirectly estimated by observing the first formant transition, which reflects changes in the first formant frequency over time. |
|---|---|

Table 2 presents the relative relationship between tenseness and the spatial configuration of the vocal tract during the production of close vowels. The indices used for this analysis are the absolute difference between the first formant frequency of the vowel (*F1*) and the first formant frequency in a neutral vocal tract configuration (*Fneu*), denoted as |*F1* - *Fneu*|, as well as the angle *θ1*.

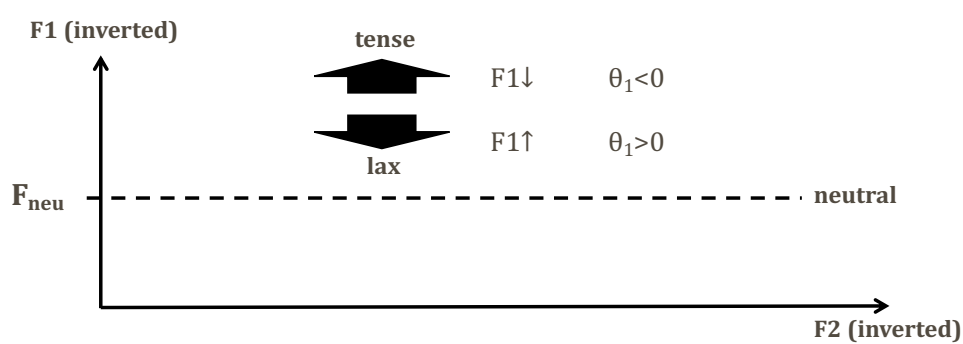

**Figure 2. Correlation Between Vowel Tenseness and First Formant Transition (For Close Vowels)**

Consider two vowels, A and B, with their respective first formant frequencies denoted as *F1A* and *F1B*. If the condition |*F1A* - *Fneu*| > |*F1B* - *Fneu*| holds, vowel A can be considered relatively more tense than vowel B. According to Chomsky and Halle (1968), tense vowels deviate further from the neutral vocal tract position than lax vowels. Given that the first formant frequency is correlated with vertical articulatory movements, vowel A, which exhibits *F1A*, is inferred to be in a relatively tenser state than vowel B, which exhibits *F1B*.

Thus, by utilizing the absolute difference |*F1* - *Fneu*| as an index, the relative tenseness of vowels can be indirectly estimated. If the time-dependent variation of |*F1* - *Fneu*| increases, it indicates an

acceleration of tenseness; if it remains unchanged, the tenseness is stable; and if it decreases, the tenseness is decelerating. Similarly, the temporal dependency of vowel tenseness can be inferred using the angle $\theta_1$.

**Table 2: Relative Relationship Between Vowel Tenseness and Vocal Tract Spatial Configuration (In the Case of Close Vowels)**

| Tenseness | Vocal Tract Spatial Configuration | $|F1-F_{neu}|$ | Angle $\vartheta_1$ |
|---|---|---|---|
| Tense | Constriction | Increase | <0 |
| Stable | No Change | No Change | =0 |
| Lax | Expansion | Decrease | >0 |

Regarding vowel tenseness, Table 2 indicates that when the angle $\theta_1$ is negative, the vowel is classified as tense, whereas when $\theta_1$ is positive, it is classified as lax. However, the relative tenseness between two vowels is not uniformly determined solely by the sign of $\theta_1$. Instead, it is suggested that the classification may depend on whether $\theta_1$ is distinctly bifurcated.

Figure 3 illustrates the relative relationship of vowel tenseness. The left panel represents a case where the two vowels have both positive and negative values of $\theta_1$, the center panel shows a case where $\theta_1$ is approximately zero and negative, and the right panel depicts a case where $\theta_1$ is approximately zero and positive. In all these cases, the relative tenseness between the two vowels can be interpreted as an indication of tense and lax distinctions.

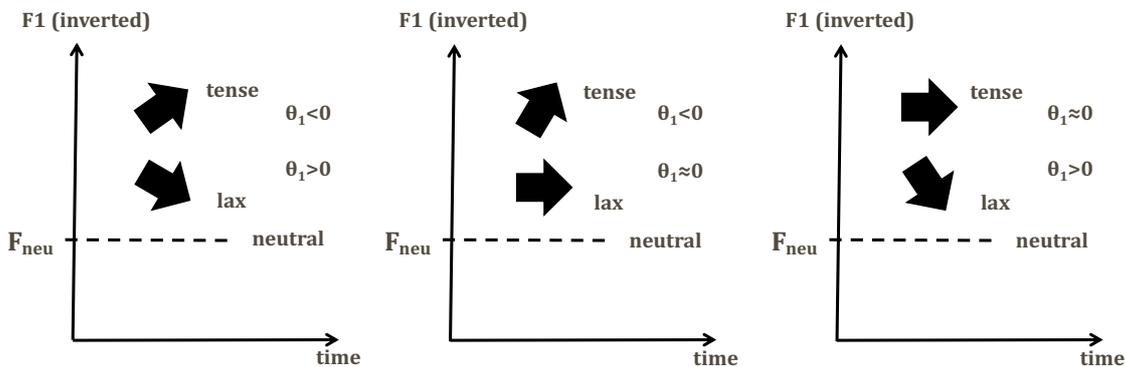

**Figure 3. Relative Relationship of Vowel Tenseness**
**(If the distribution of θ1 is bifurcated, the two vowels can be relatively classified as tense and lax.)**

## 3.2 Quantification of Vowel Tenseness at Time *t*

Ishizaki (2018c) introduced the concept of time dependence in analyzing vowel tenseness, recognizing the difficulty of precisely quantifying tenseness at a single point on the time axis. Instead of categorizing vowel tenseness in a binary manner as tense or lax, Ishizaki proposed a quantification approach using a continuous numerical representation, with the angle $\theta_1$ of the first formant frequency as the key indicator.

This study extends the discussion of vowel tenseness within a finite duration, such as vowel duration, by examining vowel tenseness at a specific time *t*. Figure 4 builds upon Figure 1, which illustrates the acoustic properties used in vowel quality research. It includes additional representations: (1) the variation in formant frequency from $Zn_{33}$ to $Zn_{66}$ within 33% to 66% of the vowel duration, and

(2) the change in formant frequency from *Zn(t)* to *Zn(t+Δt)* over a small time interval [*t, t+Δt*] within the vowel duration (displayed on the Bark scale).

One method for quantitatively analyzing formant transitions involves the use of the formant frequency angle *θn*, as proposed by Watson and Harrington (1999), which numerically represents the variation in the *n*th formant frequency over time. According to Ferguson and Kewley-Port (2007), the formant frequency angle *θn* during 33% to 66% of the vowel duration can be expressed by Equations (1)–(3). In Figure 4 and Equations (1) and (2), *d* represents the vowel duration, which is determined using the calculation shown in Equation (2) and is measured in deciseconds. The Bark scale *Z*, as defined in Equation (3), is a psychoacoustic scale derived from formant frequencies and was introduced by Traunmüller (1990).

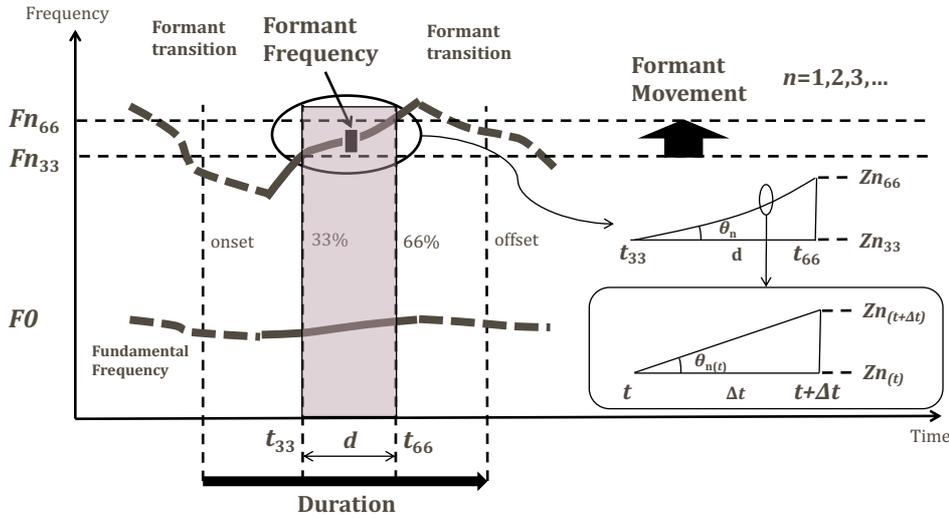

**Figure 4. Acoustic Properties Used for Analyzing Vowel Tenseness at Time *t***

$$\theta_n = arctan\left(\frac{Zn_{66} - Zn_{33}}{d}\right) \quad (1)$$

$$d = \frac{t_{66} - t_{33}}{100} \quad (2)$$

$$Z = \frac{26.81}{1 + \frac{1960}{f}} - 0.53 \quad \text{Eq. (1) (2) Ferguson \& Kewley-Port (2007)} \atop \text{Eq. (3) \quad Traunmüller (1990)} \quad (3)$$

**Mathematical model derived based on the quantitative definition of vowel tension proposed by Ishizaki (2019)**

**Vowel tenseness (duration *d*)**

$$\theta_n = arctan\left(\frac{Zn_{66} - Zn_{33}}{d}\right)$$

$$\theta_{Fn} = arctan\left(\frac{Fn_{66} - Fn_{33}}{d}\right) \quad \text{(the case n=1)}$$

**Vowel tenseness (at time *t*)  the first derivative**

$$\frac{dZn_{(t)}}{dt} = \lim_{\Delta t \to 0} \tan\theta_{n(t)} \quad (a)$$

$$\theta_{n(t)} = arctan\left(\frac{Zn_{(t+\Delta t)} - Zn_{(t)}}{\Delta t}\right) \quad \text{(the case n=1)} \quad (b)$$

**Vowel tenseness (at time *t*)  the second derivative**

$$\frac{d^2 Zn_{(t)}}{dt^2} = \frac{d}{dt}(\lim_{\Delta t \to 0} \tan\theta_{n(t)}) \quad \text{(the case n=1)}$$

To quantify vowel tenseness at a specific time *t*, the formant angle *θn* over the small time interval [*t, t+Δt*] is first defined as in Equation (b). By considering *θn(t)* as a function of time, its derivative can be obtained, serving as an indicator of vowel tenseness at time *t*. This derivative is given by Equation (a). In this analysis, the derivative is expressed as a cubic function of time; however, in a generalized form, it can be represented as an *n*-th degree function.

## 4. Results
## 4.1 Novel Phonetic Features of the English Tense /iː/ and Lax /ɪ/

Table 3 and Figure 5 present the tenseness index for the English tense and lax vowels, /iː/ and /ɪ/, as represented by the formant angle *θ1* (in radians). The tense vowel /iː/ tends to exhibit predominantly negative values, with a minimum value of -1.0640 radians, a first quartile of -0.5650 radians, a median of -0.2810 radians, a mean of -0.3087 radians, a third quartile of -0.0985 radians, and a maximum value of 0.4530 radians. In contrast, the lax vowel /ɪ/ generally exhibits positive values, with a minimum of -0.9270 radians, a first quartile of -0.0605 radians, a median of 0.5170 radians, a mean of 0.4434 radians, a third quartile of 0.9220 radians, and a maximum of 1.2770 radians.

Focusing on the median values, the tense vowel /iː/ has a median of -0.2810 radians, while the lax vowel /ɪ/ has a median of 0.5170 radians, resulting in a difference of 0.7980 radians. As illustrated in

**Table 3. Indicators of Tenseness for the English Tense and Lax Vowels /iː/ and /ɪ/, and Formant Angle $\theta_1$**

| Vowels | Min. | 1st Qu. | Median | Mean | 3rd Qu. | Max. |
|---|---|---|---|---|---|---|
| /iː/ | -1.0640 | -0.5650 | -0.2810 | -0.3087 | -0.0985 | 0.4530 |
| /ɪ/ | -0.9270 | 0.0605 | 0.5170 | 0.4434 | 0.9220 | 1.2770 |

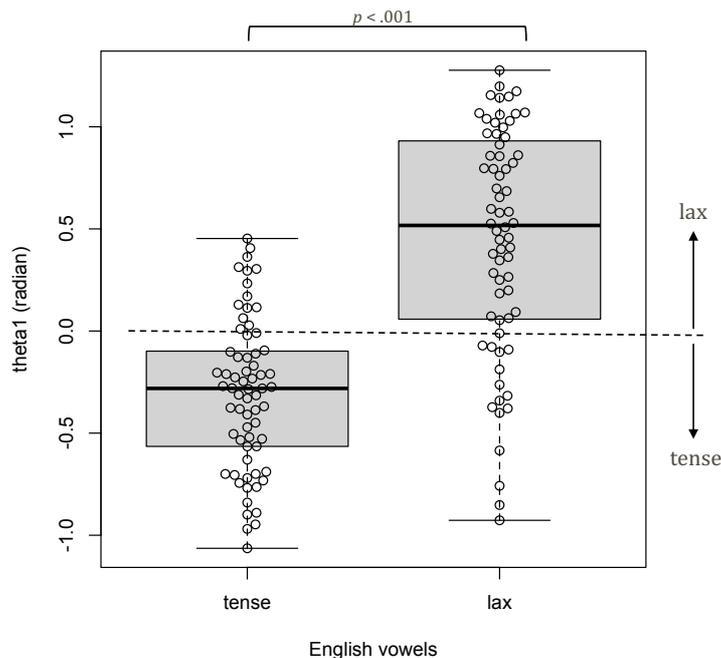

**Figure 5. Indicators of Tenseness for the English Tense and Lax Vowels /iː/ and /ɪ/, and Formant Angle $\theta_1$**

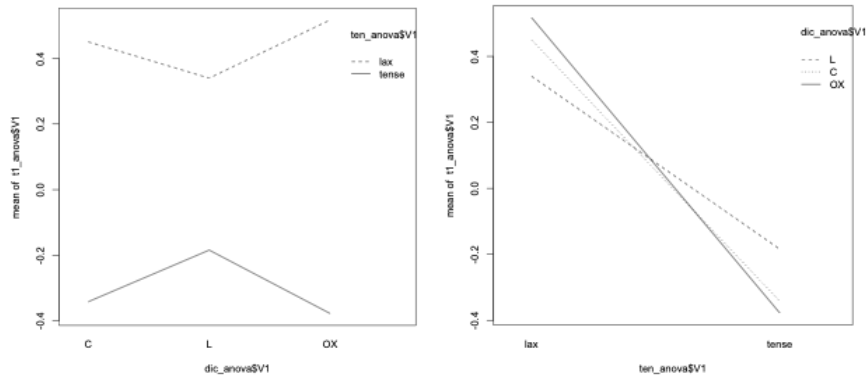

**Figure 6. Two-Way ANOVA on Formant Angle $\theta_1$
for the English Tense and Lax Vowels /iː/ and /ɪ/
Y-axis: Formant Angle $\theta_1$, X-axis: Dictionary (left) and Vowel (right)**

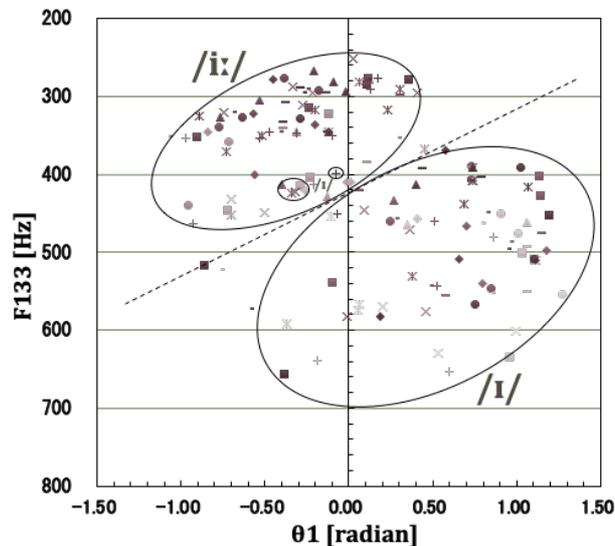

**Figure 7. Formant Angle $\theta_1$ and First Formant Frequency $F1_{33}$ for
the English Tense and Lax Vowels /iː/ and /ɪ/**

Figure 5, the distributions of the formant angle $\theta 1$ for the two vowels differ significantly. Welch's t-test confirmed a statistically significant difference in $\theta 1$ between the two conditions (tense, lax), $t(116)$ =9.24, $p < .001$.

Therefore, the formant angle $\theta 1$ for the English tense and lax vowels, /iː/ and /ɪ/, as shown in Figure 5, can be regarded as a statistically significant indicator of vowel tenseness. In other words, the "one-dimensional map of formant angle $\theta 1$" has the potential to serve as a novel phonetic feature applicable to computer-assisted language learning (CALL) in vowel pronunciation training.

Figure 6 presents the results of a two-way analysis of variance (ANOVA). The analysis revealed a significant main effect of vowel type, $F(1,129) = 84.192$, $p < .001$. However, the main effect of dictionary was not significant at the 5% level. Additionally, the interaction effect between dictionary and vowel type was not significant at the 5% level.

Figure 7 presents the indicators of tenseness for the English tense and lax vowels, /iː/ and /ɪ/: the formant angle $\theta 1$ (in radians) and the first formant frequency $F1_{33}$ (the value at 33% of the vowel duration, in Hz). Welch's t-test revealed a significant difference in $F1_{33}$ between the two conditions

(tense, lax) ($t(120) = 15.00$, $p < .001$). Therefore, both the formant angle $\theta_1$ and the first formant frequency $F1_{33}$, as shown in Figure 7, can be considered statistically significant acoustic characteristics of these vowels. In other words, the "two-dimensional map of formant angle $\theta_1$ and first formant frequency $F1_{33}$," as illustrated in Figure 7, has the potential to serve as a novel phonetic feature applicable to computer-assisted language learning (CALL) for vowel pronunciation training.

## 4.2 Novel Phonetic Features of the English Tense /uː/ and Lax /ʊ/

Table 4 and Figure 8 present the tenseness indicator for the English tense and lax vowels, /uː/ and /ʊ/, specifically the formant angle $\theta_1$ (in radians). The tense vowel /uː/ tends to exhibit predominantly negative values, with a minimum of -0.96069 radians, a first quartile value of -0.74060 radians, a median of -0.69243 radians, a mean of -0.63691 radians, a third quartile value of -0.66674 radians, and a maximum of 0.08598 radians. In contrast, the lax vowel /ʊ/ tends to exhibit predominantly positive values, with a minimum of -1.1124 radians, a first quartile value of -0.3367 radians, a median of 0.4031 radians, a mean of 0.1587 radians, a third quartile value of 0.6227 radians, and a maximum of 0.9959 radians.

Focusing on the median values, the tense vowel /uː/ has a median formant angle $\theta_1$ of -0.69243 radians, while the lax vowel /ʊ/ has a median value of 0.4031 radians, resulting in a difference of 1.09553 radians. As shown in Figure 8, the distribution of formant angle $\theta_1$ differs significantly between the two vowels. Welch's t-test revealed a statistically significant difference in $\theta_1$ between the tense and lax conditions ($t(19) = 4.0413$, $p < .001$).

**Table 4. Indicators of Tenseness for the English Tense and Lax Vowels /uː/ and /ʊ/, and Formant Angle $\theta_1$**

| Vowels | Min. | 1st Qu. | Median | Mean | 3rd Qu. | Max. |
| --- | --- | --- | --- | --- | --- | --- |
| /uː/ | -0.96069 | -0.74060 | -0.69243 | -0.63691 | -0.66674 | 0.08598 |
| /ʊ/ | -1.1124 | -0.3367 | 0.4031 | 0.1587 | 0.6227 | 0.9959 |

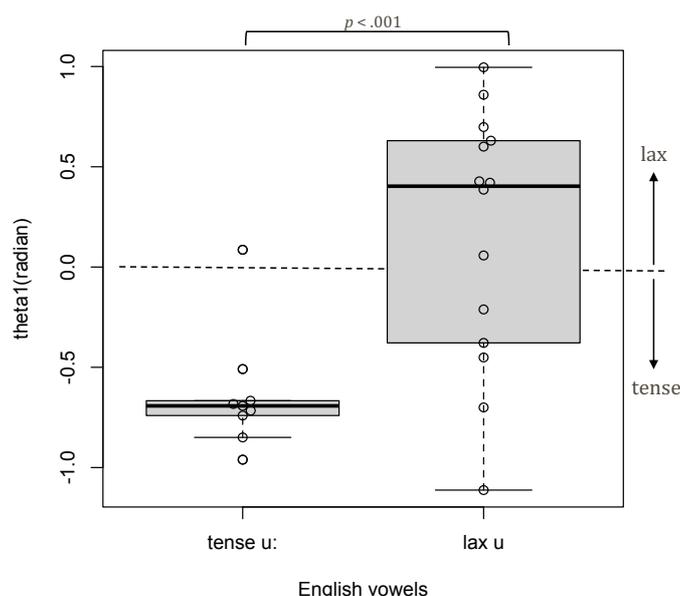

**Figure 8. Indicators of Tenseness for the English Tense and Lax Vowels /uː/ and /ʊ/, and Formant Angle $\theta_1$**

Therefore, the formant angle $θ1$ presented in Figure 8 can be regarded as a statistically significant indicator of vowel tenseness. Specifically, a one-dimensional map of formant angle $θ1$ may serve as a novel phonetic feature dataset applicable to computer-assisted language learning (CALL) and e-learning for vowel pronunciation training.

Figure 9 illustrates the relationship between the formant angle $θ1$ (in radians) and the first formant frequency at 33% of vowel duration ($F1_{33}$, in Hz) for the English tense and lax vowels /uː/ and /ʊ/. Welch's t-test indicated a statistically significant difference in $F1_{33}$ between the two vowel conditions ($p < .001$).

Thus, both the formant angle $θ1$ and $F1_{33}$, as shown in Figure 9, can be considered acoustically significant features that distinguish tense and lax vowels. Additionally, the Bark coefficient $Z1_{33}$ may be used as an alternative to $F1_{33}$. Consequently, a two-dimensional map of formant angle $θ1$ and first formant frequency $F1_{33}$, as depicted in Figure 9, may serve as a novel phonetic feature dataset for CALL and e-learning applications in vowel pronunciation training.

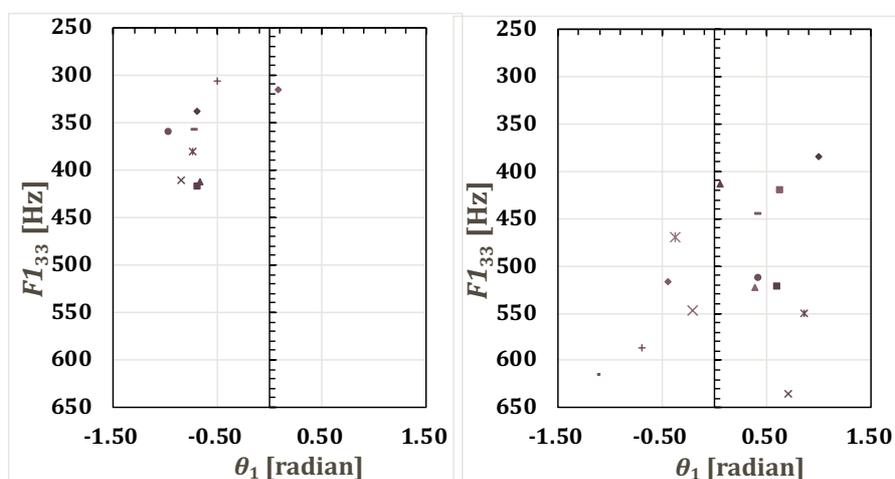

**Figure 9. Two-Dimensional Map of Formant Angle $θ_1$ and First Formant Frequency $F1_{33}$ as Indicators of Tenseness for the English Tense and Lax Vowels /uː/ and /ʊ/ (Left: /uː/, Right: /ʊ/)**

## 4.3 Novel Phonetic Features of the Japanese Tense /uR/(HL) and Lax /uR/(LH)

Table 5 and Figure 10 present the formant angle $θ1$ (in radians) as an indicator of tenseness for the Japanese vowels /uR/(HL) and /uR/(LH). The vowel /uR/(HL) tends to exhibit predominantly negative values, with a minimum of -1.03236 radians, a first quartile of -0.38656 radians, a median of -0.18350 radians, a mean of -0.15822 radians, a third quartile of 0.04787 radians, and a maximum of 0.57448 radians.

Conversely, the vowel /uR/(LH) generally exhibits positive values, with a minimum of -0.6838 radians, a first quartile of 0.1140 radians, a median of 0.2810 radians, a mean of 0.2325 radians, a third quartile of 0.4024 radians, and a maximum of 0.6918 radians.

Focusing on the median values, the vowel /uR/(HL) has a median of -0.18350 radians, while the vowel /uR/(LH) has a median of 0.2810 radians, resulting in a difference of 0.4645 radians. As shown in Figure 10, the distribution of formant angle $θ1$ differs significantly between the two vowels. Welch's t-test revealed a significant difference in $θ1$ between the HL and LH conditions ($t(54) = 4.6246, p < .001$). Therefore, the formant angle $θ1$ presented in Figure 10 can be regarded as a statistically significant indicator of vowel tenseness in Japanese vowels /uR/(HL) and /uR/(LH).

Figure 11 illustrates a two-dimensional mapping of the formant angle $\theta 1$ and pitch variation $F0_{66}-F0_{33}$ as indicators of vowel tenseness in Japanese vowels /uR/(HL) and /uR/(LH). A moderate positive correlation was observed between these variables for the Japanese vowel /uR/ ($r=0.505$), which was statistically significant ($p < .001$). The results suggest a tendency for the absolute value of the tenseness indicator θ1 to increase as the absolute difference in pitch variation becomes larger. The Japanese vowel /uR/ has been empirically demonstrated to exhibit tenseness, indicating the existence of a Tense/Lax pair.

**Table 5. Indicators of Tenseness for the Japanese Vowels /uR/(HL) and /uR/(LH), and Formant Angle $\theta_1$**

| Vowels | Min. | 1st Qu. | Median | Mean | 3rd Qu. | Max. |
|---|---|---|---|---|---|---|
| /uR/ HL | -1.03236 | -0.38656 | -0.18350 | -0.15822 | 0.04787 | 0.57448 |
| /uR/ LH | -0.6838 | 0.1140 | 0.2810 | 0.2325 | 0.4024 | 0.6918 |

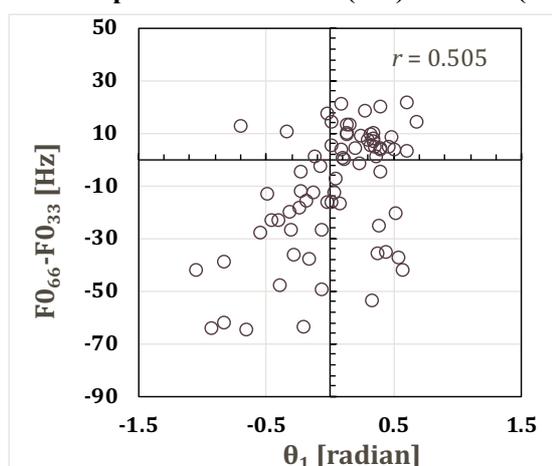

Figure 10. One-Dimensional Map of Formant Angle $\theta_1$ as an Indicator of Tenseness for the Japanese Vowels /uR/(HL) and /uR/(LH)

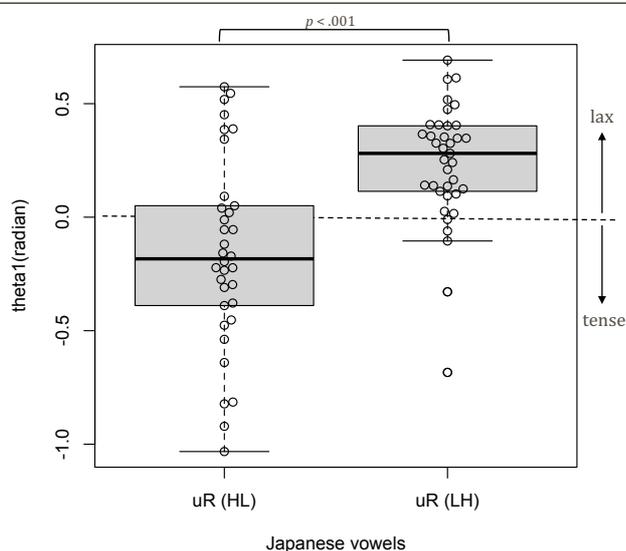

Figure 11. Formant Angle $\theta_1$ and Pitch Variation $F0_{66}$-$F0_{33}$ for the Japanese Vowels /uR/(HL) and /uR/(LH)

## 5. Discussion and Conclusion
## 5.1 Force of Tenseness as a Novel Determinant of Vowel Quality Based on a Physical Analysis

This study introduces the concept of "Force of Tenseness, *Ftense(t)*," based on a physical analysis and examines its implications for pronunciation training. The methodology employed in this study closely follows that proposed by Ishizaki (2022).

Consider a mass point of mass *m* moving along the x-axis. When subjected to a force F, the velocity *vx=dx/dt* is governed by the equation of motion given in Equation (4). In this study, the articulatory movement of the jaw and tongue when producing tense and lax vowels, such as /uː/ and /ʊ/, is modeled as the motion of a mass point along a one-dimensional x-axis (Figure 12). The acceleration *atense(t)* of the mass point *m* is defined as the second derivative of *Zn* (Equation (5)), where n=1).

$$F = m\frac{dv_x}{dt} \quad (4)$$

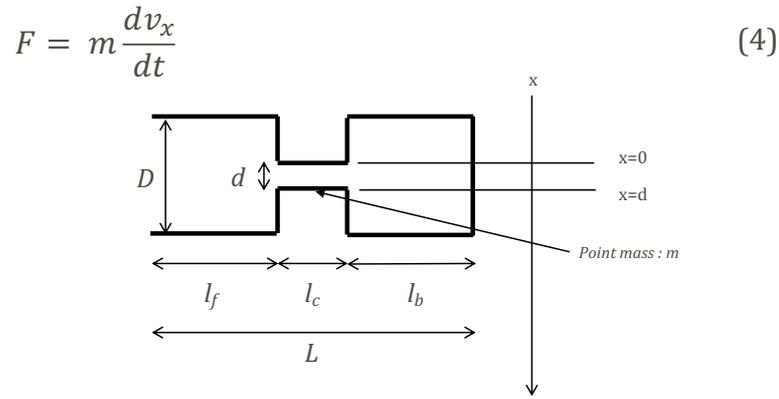

**Figure 12. Model of the Up-and-Down Motion of the Articulatory Organs During the Pronunciation of /uː/ and /ʊ/, Assuming a Mass Point with Mass *m* Moving Along the x-axis**
**The left side represents the lips, and the right side represents the glottis.**

$$a_{tense}(t) = \frac{d^2 Zn_{(t)}}{dt^2} = \frac{d}{dt}(\lim_{\Delta t \to 0} \tan \theta_{n(t)}) \quad (5)$$
$$= 6at+2b$$
(the case Zn(t) = at³ + bt² + ct + d)

$$f = \frac{c}{2\pi}\sqrt{\frac{A_c}{A_b l_b l_c}} = d\frac{c}{2\pi D\sqrt{l_b l_c}} \quad (6)$$

$$F_{tense}(t) = mka_{tense}(t) \quad (7)$$
$$= mk\frac{d^2 Zn_{(t)}}{dt^2}$$
$$= mk\frac{d}{dt}(\lim_{\Delta t \to 0} \tan \theta_{n(t)})$$
$$= mk\,(6at+2b)$$

Assuming that "muscular tension corresponds to force involvement," this study examines the force indicator Ftense(t) at time *t*. The model assumes constant acceleration without considering gravity or other external loads. The frequency of Helmholtz resonance, which corresponds to the formant frequency *F1* of the vowel, is given in Equation (6). The newly proposed vowel tenseness indicator *Ftense(t)* is defined in Equation (7). Here, *k* is a constant coefficient determined by the relationship between *Z1*, *F1*, and *d*, assuming a small range of fluctuation in *F1*.

Figure 13 presents the calculated values of *atense(t)* for the English tense vowel /uː/ (e.g., *hoop*) and the Japanese vowel /uR/(HL) (e.g., *gūsū*). Assuming that the mass *m* and coefficient *k* are identical for speakers producing both English and Japanese vowels, it is possible to estimate the behavior of the "Force of Tenseness, *Ftense(t)*" by observing *atense(t)*. The results indicate that *atense(t)* in English vowels is approximately an order of magnitude higher than that in Japanese vowels. This suggests that the "Force of Tenseness, *Ftense(t)*" for English tense vowel /uː/ is significantly greater than that for the Japanese vowel /uR/(HL). Thus, this study demonstrates the theoretical feasibility of indirectly calculating the "Force of Tenseness, *Ftense(t)*" using first formant frequency and comparing vowels across languages.

In the field of linguistics, three key factors—tongue height, tongue frontness-backness, and lip position—are widely acknowledged as the primary determinants of vowel quality in English, all of which are critical for pronunciation instruction. These factors are considered fundamental by both researchers and educators in the field of English language education. This study suggests the potential inclusion of a fourth determinant: "Ftense (t) = Force of Tenseness," which is supported by a physical analysis.

The degree of vowel tenseness may be influenced by variables such as the velocity and acceleration of the tongue's vertical movement within the oral cavity. Drawing on the physical relationship between acceleration and force, the analysis proposes that specific dependencies may exist for the "Force of Tenseness" in vowel production. This study introduces the possibility that this "force" element could serve as a novel determinant of vowel quality, arguing that it warrants further investigation as a crucial area for future research.

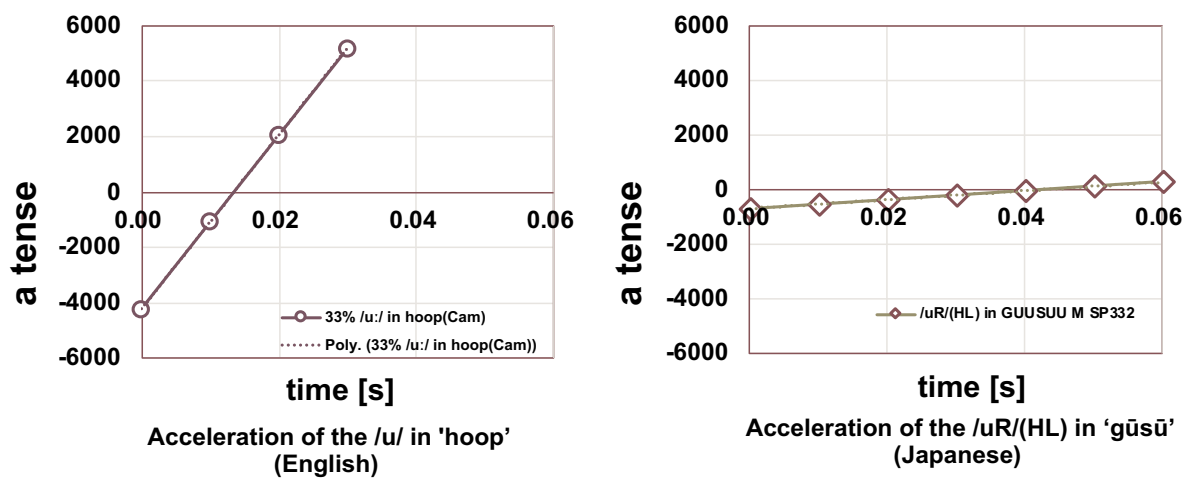

**Figure 13: Acceleration of Tenseness, *atense(t)*,
in the English Tense Vowel /uː/ and the Japanese Tense Vowel /uR/(HL)**

## 5.2 A Lagrangian Model for Force of Tenseness

In the previous model, the x-axis was applied to the tube model. However, constructing a more generalized model that incorporates multiple axes and accounts for the rotational motion of the jaw will be essential for future analyses of "Force of Tenseness." To advance this framework, this section introduces the Euler-Lagrange equation as a basis for modeling "Force of Tenseness" in a complex system. By utilizing this generalized equation, the forces generated during vowel production can be systematically modeled, enabling energy analysis and the estimation of "Force of Tenseness."

To establish this foundation, we adopt a simplified model in which the x-axis is oriented vertically downward, with $x=0$ representing the position of the palate and $x=d$ denoting the distance from the palate to the surface of the tongue. The y-axis is assumed to be oriented horizontally to the right, with $y=0$ serving as the reference point for the anterior-posterior position of the tongue. While the mathematical formulation for y-axis dynamics will be explored in future research, the present model considers only forces along the x-axis. However, our long-term goal is to develop a generalized model that integrates forces along the y-axis as well as the rotational dynamics of the jaw.

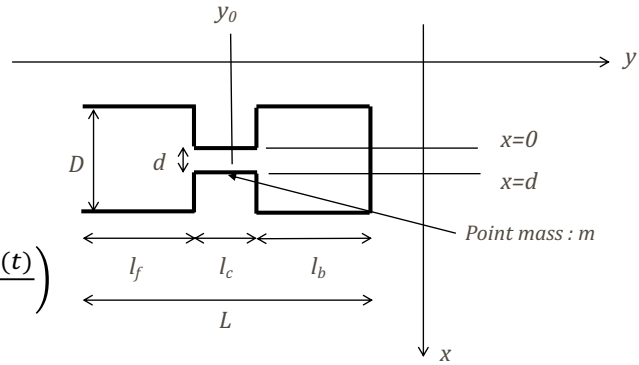

$$\frac{d}{dt}\left(\frac{\partial L}{\partial \dot{q}_i}\right) - \frac{\partial L}{\partial q_i} = 0$$

$$\frac{\partial L}{\partial \dot{x}} = mk \lim_{\Delta t \to 0} \tan \theta_{n(t)}$$

$$\theta_{n(t)} = \arctan\left(\frac{Zn_{(t+\Delta t)} - Zn_{(t)}}{\Delta t}\right)$$

$$L = \frac{1}{2} mk \sum_{i=0}^{n} c_i t^i$$

$$F_x = mk \frac{d}{dt}\left(\lim_{\Delta t \to 0} \tan \theta_{n(t)}\right)$$

based on the observed relationship between $x$ and $Z_n$.

$$L = \frac{1}{2} m v_y^2 - \frac{1}{2} p (y - y_0)^2$$

$$F_y = -p (y - y_0)$$

For a simple harmonic oscillator, $p$ represents the spring constant, and $y_0$ denotes the equilibrium position. (assumed)

**Figure and Table 1. Generalized Lagrangian Model for Estimating Energy and Forces in Vowel Production.**

This paper not only provides an explanation of Ishizaki (2019) and Ishizaki (2023), which are written in Japanese, but also introduces a new energy model related to vowel tenseness using a Lagrangian approach.

## Acknowledgment

This research was supported by JSPS Kakenhi (JP23K12151, Principal Investigator: Tatsuya Ishizaki) and JSPS Kakenhi (JP21K19976, Principal Investigator: Tatsuya Ishizaki). I would like to express my sincere gratitude for the support.

## References

Balbonas, Dainius and Gintautas Daunys (2015) Movement of formants of vowels in Lithuanian language. *Elektronika ir Elektrotechnika,* 79(7), 15-18.
Chiba, Tsutomu and Masato Kajiyama (1941) *The vowel: Its nature and structure*. Tokyo: Tokyo-Kaiseikan.
Chomsky, Noam and Morris Halle (1968) *The sound pattern of English*. New York: Harper & Low.
Fant, Gunnar (1960) *Acoustic theory of speech production*. The Hague: Mouton.
Ferguson, Sarah H. and Diane Kewley-Port (2007) Talker differences in clear and conversational speech: Acoustic characteristics of vowels. *Journal of Speech, Language, and Hearing Research,* 50, 1241-1255.
Hillenbrand, James, Laura A. Getty, Michael J. Clark and Kimberlee Wheeler (1995) Acoustic characteristics of American English vowels. *The Journal of the Acoustical Society of America,* 97(5): 3099-3111.
Ishizaki, Tatsuya (2015) A study on English vowel pronunciation instruction methods focusing on formant movement. Presentation at the 19th Annual Conference of the Foreign Language Education Society, Tokyo University of Foreign Studies, November 29, 2015.
Ishizaki, Tatsuya (2016) A study on English vowel pronunciation instruction methods focusing on formant movement. *Journal of linguistic science, Tohoku University*, 20: 1-12. Sendai: Tohoku University.
Ishizaki, Tatsuya (2018a) Tenseness and Formant Movement in Japanese Vowels. *Tohoku University Linguistics Journal*, 27: 69-80. Sendai: Tohoku University.
Ishizaki, Tatsuya (2018b) Accent and Tenseness in Japanese Vowels. *Journal of linguistic science, Tohoku University*, 22: 1-12. Sendai: Tohoku University.
Ishizaki, Tatsuya (2018c) Quantification of vowel tenseness and an attempt at quantitative evaluation in pronunciation education. Presentation at the 22nd Annual Conference of Japan Association of Foreign Language Education, Tokyo University of Foreign Studies, December 15, 2018.
Ishizaki, Tatsuya (2019) Quantification of vowel tenseness at time t – Focusing on the time dependence of formant frequencies. *Tohoku University Linguistics Journal*, 28, 45-58.
Ishizaki, Tatsuya (2020) Quantification of vowel tenseness using F0 and F1 and a study on pronunciation methods – Focusing on the teaching of tense-lax vowel pairs /iː/ and /ɪ/. *Japan Association of Foreign Language Education bulletin*, 23, 147-159.
Ishizaki, Tatsuya (2021) Control of vowel tenseness with F0 variation and the development of a pronunciation method: Pronunciation of tense/lax vowels /iː/ and /ɪ/ for Japanese learners of English. *Japan Association of Foreign Language Education bulletin*, 24, 99-112.
Ishizaki, Tatsuya (2022) Vowel Tenseness θ1 and Formant Frequency F1 of Tense /iː/ and Lax /ɪ/. *Japan Association of Foreign Language Education bulletin*, 25, 188-205.
Ishizaki, Tatsuya (2023) Pronunciation Teaching Methods of English Tense/Lax Vowels, /uː/ and /ʊ/ Based on Vowel Tenseness, θ1 and Ftense(t): A Quantitative Study of Japanese Tense and Lax Vowels, /uR/(HL) and /uR/(LH). *Japan Association of Foreign Language Education bulletin*, 26, 1-19.
Iverson, Paul and Bronwen G. Evans (2007) Learning English vowels with different first-language vowel systems: Perception of formant targets, formant movement, and duration. *The Journal of the Acoustical Society of America,* 122(5): 2842-2854.
Jacewicz, Ewa and Robert A. Fox (2013) Cross-dialectal differences in dynamic formant patterns in American English vowels. *Vowel inherent spectral change*, 177-198. Berlin, Heidelberg: Springer.
Jakobson, Roman, C. Gunnar Fant, and Morris Halle (1951) *Preliminaries to speech analysis: The distinctive features and their correlates.* Cambridge: The MIT Press.

McDougall, Kirsty and Francis Nolan (2007) Discrimination of speakers using the formant dynamics of /uː/ in British English. *Proceedings of the 16th International congress of phonetic sciences,* 1825-1828.
Schwartz, Geoffrey (2015) Vowel dynamics for Polish learners of English. *Teaching and researching the pronunciation of English,* 205-217. Cham: Springer.
Slifka, Janet (2003) Tense/lax vowel classification using dynamic spectral cues. *Proceedings of 15th International conference of phonetic sciences,* 921-924.
Traunmüller, Hartmut (1990) Analytical expressions for the tonotopic sensory scale. *The Journal of the Acoustical Society of America*, 88(1): 97-100.
Watson, Catherine I. and Jonathan Harrington (1999) Acoustic evidence for dynamic formant trajectories in Australian English vowels. *The Journal of the Acoustical Society of America,* 106(1): 458-468.
Wrede, Britta, Gernot A. Fink and Gerhard Sagerer (2000) Influence of duration on static and dynamic properties of German vowels in spontaneous speech. *Sixth International conference on spoken language processing,* 1: 82-85.
Yang, Jing and Robert A. Fox (2013) Acoustic development of vowel production in American English children. *Interspeech,* 1263-1267.
Yang, Jing, and Robert A. Fox (2017) Acoustic development of vowel production in native Mandarin-speaking children. *Journal of the International Phonetic Association*, 1-19.

**Online Dictionaries:**

*CAMBRIDGE DICTIONARY*. https://dictionary.cambridge.org/ [accessed November 2021].
*LONGMAN ENGLISH DICTIONARIES.* https://www.ldoceonline.com/dictionary/ [accessed November 2021].
*OXFORD LEARNER'S DICTIONARIES.* https://www.oxfordlearnersdictionaries.com/ [accessed November 2021].

**Online Databases:**

*National Institute of Informatics Speech Resource Consortium* (NII SRC). http://research.nii.ac.jp/src/index.html[accessed March 2022].